\pdfoutput=1

\documentclass[11pt]{article}

\usepackage[]{naacl2021}

\usepackage{times}
\usepackage{latexsym}

\usepackage[T1]{fontenc}

\usepackage[utf8]{inputenc}

\usepackage{microtype}
\usepackage{hyperref}
\usepackage{float}
\usepackage{subcaption}
\usepackage{graphicx}  
\usepackage{todonotes}
\usepackage{enumitem}
\usepackage{comment}
\usepackage{multirow}

\title{Towards Robust Neural Retrieval Models with Synthetic Pre-Training}

\author{Revanth Gangi Reddy$^1$, Vikas Yadav$^2$, Md Arafat Sultan$^2$, Martin Franz$^2$,\\ \textbf{Vittorio Castelli$^2$, Heng Ji$^1$, Avirup Sil$^2$} \\
  $^1$University of Illinois at Urbana-Champaign 
   $^2$IBM Research AI, New York \\
   \texttt{\{revanth3,hengji\}@illinois.edu, arafat.sultan@ibm.com,}\\
   \texttt{\{vikasy,franzm,vittorio,avi\}@us.ibm.com} }

\begin{document}
\maketitle

\begin{abstract}
Recent work has shown that commonly available machine reading comprehension (MRC) datasets can be 
used to train high-performance neural information retrieval (IR) systems.
However, the evaluation of neural IR has so far been limited to standard supervised learning settings, where they have outperformed traditional term matching baselines.
We conduct in-domain and out-of-domain evaluations of neural IR, and seek to improve its robustness across different 
scenarios, including zero-shot settings.
We show that synthetic training examples generated using a sequence-to-sequence generator can be 
effective towards this goal: in our experiments, pre-training with synthetic examples improves retrieval performance in both in-domain and out-of-domain evaluation on five different test sets.

\end{abstract}

\section{Introduction}


\begin{figure*}[ht]
     \begin{subfigure}[c]{0.265\textwidth}
    \centering
     \includegraphics[scale=0.74]{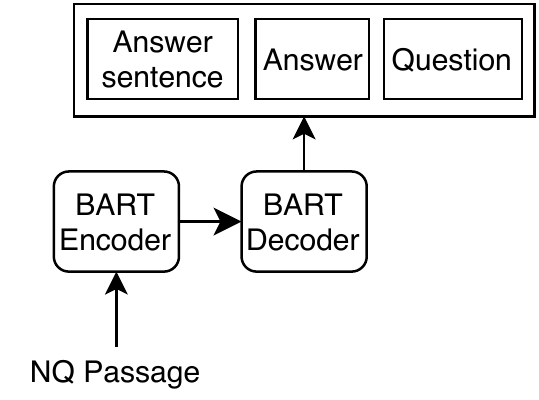}
       \caption{\vspace{-0.7475cm}}
       \label{fig:syn_train}
     \end{subfigure}
     \hfill
     \begin{subfigure}[c]{0.32\textwidth}
     \centering
     \includegraphics[scale=0.74]{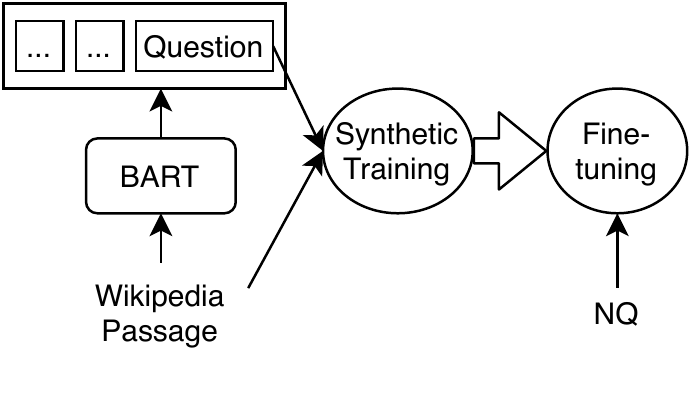}
       \caption{\vspace{-0.7675cm}}
       \label{fig:syn_inference}
     \end{subfigure}
     \begin{subfigure}[c]{0.40\textwidth}
     \centering
     \small
        \begin{tabular}{p{5.8cm}}
            \textbf{History of Tanzania} The African Great Lakes nation of Tanzania dates formally from \textcolor[rgb]{0,0,.7}{\textbf{\textit{1964}}}, when it was formed out of the union of the much larger mainland territory of Tanganyika and the coastal archipelago of Zanzibar. The former was a colony and part of German East Africa from the 1880s to 1919, when, under the League of Nations, it became a  \textcolor[rgb]{.5,0,0}{\textbf{\textit{British mandate}}}. It served as a military outpost ... \\
        \hline
            \textcolor[rgb]{0,0,.7}{\textit{when did tanzania became a country in africa?}} \\
            \textcolor[rgb]{.5,0,0}{\textit{who owned zanzibar and tanganyika before they were independent?}} \\
        \end{tabular}
       \caption{}
       \label{fig:syn_ex}
     \end{subfigure}
     \caption{Proposed IR training pipeline along with a synthetic example. (a) A BART encoder-decoder LM is fine-tuned on NQ for QA example generation. (b) Synthetic examples generated from Wikipedia passages are used to pre-train the neural IR model before fine-tuning on NQ. (c) Two synthetic questions output by our generator from the depicted Wikipedia passage, with corresponding answers highlighted in the text. }
     \label{fig:syn_gen}
\end{figure*}

Information retrieval (IR) aligns search queries to relevant documents or passages in large document collections.
Traditional approaches such as TF-IDF \cite{10.5555/576628} and BM25 \cite{robertson2009probabilistic} rely on simple lexical matching for query-passage alignment.
In contrast, neural IR encodes passages and questions into continuous vector representations, enabling deeper semantic matching.

Modern neural IR methods \cite{lee2019latent, chang2019pre} typically employ a dual encoder architecture \cite{bromley1993signature}, where separate pre-trained masked language models (MLMs) \cite{devlin2019bert} are fine-tuned for question and passage encoding.
\newcite{karpukhin2020dense} show that effective weak supervision for this fine-tuning step can be derived from existing machine reading comprehension (MRC) datasets \cite{kwiatkowski2019natural, joshi2017triviaqa}. 
Their Dense Passage Retriever (DPR) model demonstrates state-of-the-art (SOTA) retrieval performance on multiple Wikipedia datasets \cite{kwiatkowski2019natural, joshi2017triviaqa, berant2013semantic, baudivs2015modeling}, with  9--19\% performance improvement over term matching baselines like BM25.
DPR has also been shown to outperform other neural methods that employ sophisticated IR-specific pre-training schemes such as the Inverse Cloze Task \cite{lee2019latent} or latent learning of the retriever during MLM pre-training \cite{guu2020realm}.
Recent open domain QA systems \cite{lewis2020retrieval, izacard2020leveraging} have successfully used DPR to obtain SOTA performance on end-to-end QA.


However, despite this remarkable success of 
neural IR in standard supervised learning settings---where the training and test instances are sampled from very similar distributions---it is not clear if and to what extent such an approach would generalize to more practical zero-shot settings.
In this paper, we shed light on this question via an empirical study of DPR zero-shot performance 
on multiple datasets.
Our experiments on four out-of-domain datasets show that the advantage over BM25 continues to hold in near domains (datasets with  Wikipedia articles comprising the retrieval corpus), but not in a far domain (biomedical text).

Inspired by the success of synthetic training examples in 
MRC \cite{shakeri-etal-2020-end, zhang-etal-2020-multi-stage, sultan2020importance}, we further investigate if a similar approach can be useful for IR, including zero-shot application settings.
Concretely, we train a sequence-to-sequence generator using existing MRC data and use it to generate synthetic question-answer pairs from source domain passages.
Following the procedure of \citet{karpukhin2020dense}, we then create IR pre-training examples from these synthetic MRC examples.
We observe that pre-training on the generated examples (before fine-tuning with human annotated examples) improves the robustness of DPR: performance consistently improves across all our in-domain and out-of-domain test sets.
Importantly, the gap with BM25 in far domain evaluation is also significantly reduced.



The main contributions of this paper are:
\setlist[itemize]{noitemsep,nolistsep}
\begin{itemize}
    \item We conduct an 
    empirical evaluation of SOTA neural IR on 
    multiple in-domain and out-of-domain test sets and show how its effectiveness varies across different test conditions.
    \item We show that synthetic pre-training significantly improves the robustness of neural IR.
    \item We achieve new SOTA performance in neural IR on five different datasets, including zero-shot settings.
\end{itemize}

\section{Synthetic Pre-Training of IR}

Let $c$ be a corpus and $d \in c$ be a document; an IR example consists of a question $q$ and a passage $p$ in $d$ such that $p$ contains an answer $a$ to $q$.
Let $s$ in $p$ be the sentence that contains $a$.

We train an example generator that selects first a candidate sentence $s$ from an input $p$, then a candidate answer $a$ in $s$, and finally generates a corresponding question $q$.
To achieve this, we fine-tune BART \cite{lewis-etal-2020-bart}, a pre-trained encoder-decoder language model, to generate an ordered triple $(s,a,q)$ from $p$.
Labeled $(p, s, a, q)$ tuples for training are collected from Natural Questions (NQ) \cite{kwiatkowski2019natural}, an existing MRC dataset over Wikipedia articles.
In practice, we approximate the selection of the answer sentence $s$ by generating its first and last word.
Finally, $(q,p)$ is retained as a synthetic IR example.
Our method closely resembles that of \citet{shakeri-etal-2020-end}, except that we additionally perform an explicit selection of $s$.

In our experiments, we generate synthetic training examples from Wikipedia articles.
Following \cite{sultan2020importance}, we use top-$p$ top-$k$ sampling \cite{HoltzmanBDFC20} to promote diversity in the generated examples.
Training and inference for the synthetic example generator are depicted in Figures~\ref{fig:syn_train} and \ref{fig:syn_inference} respectively.
Figure~\ref{fig:syn_ex} shows two example questions 
output by the generator from a Wikipedia passage.

For each generated question $q$, we obtain a negative example by sampling a passage from corresponding BM25 retrievals that does not contain the generated answer $a$.
Following \citet{karpukhin2020dense}, we use in-batch negatives while training the DPR model. 
After pre-training with synthetic data, we finally fine-tune the model on IR examples derived from existing MRC data. 
We name this augmented DPR model \textit{AugDPR}. We refer the reader to \cite{karpukhin2020dense} for a more detailed description of the DPR training process.

\section{Experimental Setup}

\begin{table*}[ht]
    \centering
    \small
    \begin{tabular}{c|p{8.7cm}|p{4.0cm}}
    \multicolumn{1}{c|}{\textbf{Dataset}} &
    \multicolumn{1}{c|}{\textbf{Question}}     &  \multicolumn{1}{c}{\textbf{Answers}}   \\
    \hline
    \multirow{2}{*}{NQ} & 
    what does hp mean in war and order
    
    who was named african footballer of the year 2014
    & 
    [`hit points or health points']
    
    [`Yaya Touré']\\
    \hline
    \multirow{2}{*}{TriviaQA} & Who was the man behind The Chipmunks? 
    
    On a standard dartboard, which number lies between 12 and 20? 
    & 
    [`David Seville']
    
    [`five', `5']\\
    \hline
    \multirow{2}{*}{WebQuestions} &
    who was richard nixon married to? 
    
    what highschool did harper lee go to?
    &
    [`Pat Nixon']
    
    [`Monroe County High School']\\
    \hline
    \multirow{2}{*}{WikiMovies} &
    what does Tobe Hooper appear in?
    
    Mick Davis directed which movies?
    &
    [`Body Bags']
    
    [`The Match']
    \\
    
    
    \hline \multirow{2}{*}{BioASQ} &
    Which antiepileptic drug is most strongly associated with spina bifida? 
    
    Which receptor is inhibited by bimagrumab?
    & 
    [`Valproate']
    
    [`activin type II receptors']\\
        
    \end{tabular}
    \caption{Sample questions from each of the datasets used in evaluation.}
    \label{tab:Sample_ques}
\end{table*}

\subsection{Datasets}

We briefly describe our datasets in this section.
Example QA pairs from each are shown in Table~\ref{tab:Sample_ques}.
\newline

\vspace{-.25cm}
\noindent\textbf{Training and In-Domain Evaluation.}
We train our systems on Natural Questions (NQ) \cite{kwiatkowski2019natural}, a dataset 
with questions from Google's search log and  their human-annotated answers from Wikipedia articles. 
\citet{lewis2020question} report that 30\% of the NQ test set questions have near-duplicate paraphrases in the training set and 60--70\% of the test answers are also present in the training set. 
For this reason, in addition to the entire NQ test set, we also use the non-overlapping subsets released by \citet{lewis2020question} for in-domain evaluation.\\

\vspace{-.25cm}
\noindent\textbf{Near Domain Evaluation.}
For zero-shot near domain evaluation, where Wikipedia articles constitute the retrieval corpora,
we use the test sets of three existing datasets.

\noindent \textbf{\textit{TriviaQA}} \cite{joshi2017triviaqa} contains questions collected from trivia and quiz league websites, which are created by Trivia enthusiasts. 

\noindent \textbf{\textit{WebQuestions (WQ)}} \cite{berant2013semantic} consists of questions obtained using the Google Suggest API, with answers selected from entities in Freebase by AMT workers.

\noindent \textbf{\textit{WikiMovies}} \cite{miller2016key} contains question-answer pairs on movies, built using the OMDb and MovieLens databases. We use the test split adopted in~\cite{chen2017reading}.\\

\vspace{-.25cm}
\noindent\textbf{Far Domain Evaluation.}
For zero-shot far domain evaluation, we use a biomedical dataset.


\noindent \textbf{\textit{BioASQ}} \cite{tsatsaronis2015overview} is a competition\footnote{\href{http://bioasq.org/participate/challenges}{http://bioasq.org/participate/challenges}} on large-scale biomedical semantic indexing and QA.
We evaluate on all factoid question-answer pairs from the training and test sets of task 8B.
\newline

\vspace{-.25cm}
\noindent \textbf{Retrieval Corpora.} For  NQ, TriviaQA, WebQuestions and WikiMovies,  we use the  21M Wikipedia passages from \citet{karpukhin2020dense} as the retrieval corpus.
For BioASQ, we take the abstracts of PubMed articles from task 8A and split into passages of up to 120 words (preserving sentence boundaries). 
Table \ref{tab:datasets} shows the sizes of the corpora and the number of questions in each test set.

\begin{table}[t]
\small
\centering
\begin{tabular}{l|c|c|c}
\multicolumn{1}{c|}{Dataset} & Domain & Passages & Questions \\
\hline
NQ & Wikipedia  & 21.0M & 3,610\\
TriviaQA & Wikipedia  & 21.0M & 11,313\\
WebQuestions & Wikipedia  & 21.0M & 2,032 \\
WikiMovies & Wikipedia   & 21.0M  & 9,952\\
BioASQ & Biomedical  & 37.4M & 1092\\
\end{tabular}
\caption{Statistics of the retrieval corpus and test sets for different datasets used for evaluating the information retrieval models.}
\label{tab:datasets}
\end{table}

\begin{table*}[h]
\centering
\begin{tabular}{l|ccc|ccc|ccc}
\multicolumn{1}{c|}{Model} & \multicolumn{3}{c|}{Total} & \multicolumn{3}{c|}{No answer overlap} & \multicolumn{3}{c}{No question overlap}  \\ 
\hline
 & Top-1 & Top-10 & Top-20 & Top-1 & Top-10 & Top-20 & Top-1 & Top-10 & Top-20 \\ 

BM25 & 30.5 & 54.5 & 62.5 & 26.4 & 47.1 & 54.7 & 31.0 & 52.1 & 59.8 \\
DPR & 46.3 & 74.9 & 80.1 & 32.2 & 62.2 & 68.7 & 37.4 & 68.5 & 75.3 \\
AugDPR & \textbf{46.8} & \textbf{76.0} & \textbf{80.8} & \textbf{36.0} & \textbf{65.0} & \textbf{70.8} & \textbf{41.4} & \textbf{70.8} & \textbf{76.6} \\
\end{tabular}
\caption{NQ top-$k$ retrieval results. Performance improves across the board with synthetic pre-training (AugDPR), but more on the non-overlapping subsets of \citet{lewis2020question}.} 
\label{tab:NQ_test}
\end{table*}

\begin{table*}[ht]
\centering
\begin{tabular}{l|cc|cc|cc|cc}
 & \multicolumn{6}{c|}{Near Domains} & \multicolumn{2}{c}{Far Domain} \\
\cline{2-9}
\multicolumn{1}{c|}{Model} & \multicolumn{2}{c}{TriviaQA} & \multicolumn{2}{c}{WebQuestions} & \multicolumn{2}{c|}{WikiMovies} & \multicolumn{2}{c}{BioASQ}  \\ 
\hline
 & Top-20 & Top-100 & Top-20 & Top-100 & Top-20 & Top-100 & Top-20 & Top-100 \\ 

BM25 & 66.9 & 76.7 & 55.0 & 71.1 & 54.0 & 69.3 & \textbf{42.1}  & 50.5 \\
DPR & 69.0 & 78.7 & 63.0 & 78.3 & 69.8 & 78.1 & 34.7  & 46.9 \\
AugDPR & \textbf{72.2} & \textbf{81.1} & \textbf{71.1} & \textbf{80.8} & \textbf{72.5} & \textbf{80.7} & 41.4 & \textbf{52.4} \\
\hline
Supervised & 79.4 & 85.0 & 73.2 & 81.4 & - & - & - & - \\
\end{tabular}
\caption{Zero-shot neural retrieval accuracy improves with synthetic pre-training (AugDPR) in all out-of-domain test settings. However, BM25 remains a strong baseline on the far domain dataset of BioASQ. Numbers for the supervised models are taken from \cite{karpukhin2020dense}.} 
\label{tab:zero_shot}
\end{table*}

\subsection{Setup}

\noindent \textbf{Training.} We train the synthetic example generator using the \textit{(question, passage, answer)} triples from NQ.
Then we randomly sample 2M passages from our Wikipedia retrieval corpus and generate around 4 synthetic questions per passage.
We experimented with more passages but did not see any improvements on the NQ dev set. Please see appendix for more details on these experiments.
For top-$p$ top-$k$ sampling, we use $p=0.95$ and $k=10$.

During synthetic pre-training of DPR, for each of the 2M passages, we randomly choose one of its synthetic questions at each epoch to create a synthetic training example.
After synthetic pre-training, we fine-tune DPR on NQ to get the AugDPR model. We refer the reader to the appendix for hyperparameter details. 
\newline

\vspace{-.25cm}
\noindent \textbf{Baselines and Metrics.} We use BM25 as our term matching baseline. As a stronger neural baseline, we use the DPR-single model 
trained on NQ and released\footnote{\href{https://github.com/facebookresearch/DPR}{https://github.com/facebookresearch/DPR}} by \citet{karpukhin2020dense}. Both DPR and AugDPR use \textit{BERT-base-uncased} for question and passage encoding. As in
\cite{karpukhin2020dense}, 
our evaluation metric is top-$k$ retrieval accuracy, the percentage of questions with at least one answer in the top $k$ retrieved passages.

\section{Results and Discussion}


Table \ref{tab:NQ_test} shows NQ results on the entire test set as well as on the two subsets released by \citet{lewis2020question}.
Synthetic pre-training yields larger gains on the non-overlapping splits, with up to 4-point improvement in top-1 retrieval accuracy.

To assess the cross-domain utility of AugDPR, we evaluate it zero-shot on both near and far domain test sets.
Table \ref{tab:zero_shot} shows the results. 
For comparison, we also show numbers for supervised models reported by \citet{karpukhin2020dense} on TriviaQA and WebQuestions where the DPR model was trained directly on the training splits of these datasets.
For the near domain datasets, we observe that both DPR and AugDPR outperform BM25 by a sizable margin; additionally, AugDPR consistently outperforms DPR.
Furthermore, performance of AugDPR on WebQuestions is comparable to the supervised model.
On the far domain, however, we observe BM25 to be a very strong baseline, with clearly better scores than DPR.
The synthetic pre-training of AugDPR reduces this gap considerably, 
resulting in a slightly lower top-20 score but a 2-point gain in top-100 score over BM25. 

To investigate the relative underperformance of neural IR on BioASQ, we take a closer look at the vocabularies of the two domains of Wikipedia articles and biomedical literature.
Following \citet{gururangan-etal-2020-dont}, we compute the overlap between the 10k most frequent tokens (excluding stop words) in the two domains, represented by 3M randomly sampled passages from each.
We observe a vocabulary overlap of only 17\%, which shows that the two domains are considerably different in terminology, explaining in part the performance drop in our neural models.
Based on these results, we also believe that performance of neural IR in distant target domains can be significantly improved via pre-training on synthetic examples that are generated from raw text in the target domain.
We plan to explore this idea in future work.

We also examine the lexical overlap between the questions and their passages, since a high overlap would favor term matching methods like BM25.
We find that the coverage of the question tokens in the respective gold passages is indeed higher in BioASQ: 72.1\%,  compared to 58.6\% and 63.0\% in NQ and TriviaQA, respectively.

\citet{karpukhin2020dense} report that DPR fine-tuning takes around a day on eight 32GB GPUs, 
which is a notable improvement over more computationally intensive pre-training approaches like \cite{lee2019latent, guu2020realm}. Our synthetic pre-training takes around two days on four 32GB GPUs, which is comparable with fine-tuning in terms of computational overhead.

\section{Conclusions and Future Work}
We have shown that pre-training SOTA neural IR models with a large amount of synthetic examples improves robustness to degradation in zero-shot settings.  Our experiments show consistent performance gains in five in-domain and out-domain test sets, even in far target domains with significant vocabulary mismatch with the training set.
Future work will explore zero-shot domain adaptation of neural IR systems with synthetic examples generated from target domain raw text.  


\bibliography{anthology,custom}

\begin{thebibliography}{24}
\expandafter\ifx\csname natexlab\endcsname\relax\def\natexlab#1{#1}\fi

\bibitem[{Baudi{\v{s}} and {\v{S}}ediv{\`y}(2015)}]{baudivs2015modeling}
Petr Baudi{\v{s}} and Jan {\v{S}}ediv{\`y}. 2015.
\newblock Modeling of the question answering task in the yodaqa system.
\newblock In \emph{International Conference of the Cross-Language Evaluation
  Forum for European Languages}, pages 222--228. Springer.

\bibitem[{Berant et~al.(2013)Berant, Chou, Frostig, and
  Liang}]{berant2013semantic}
Jonathan Berant, Andrew Chou, Roy Frostig, and Percy Liang. 2013.
\newblock Semantic parsing on freebase from question-answer pairs.
\newblock In \emph{Proceedings of the 2013 Conference on Empirical Methods in
  Natural Language Processing}, pages 1533--1544.

\bibitem[{Bromley et~al.(1993)Bromley, Bentz, Bottou, Guyon, LeCun, Moore,
  S{\"a}ckinger, and Shah}]{bromley1993signature}
Jane Bromley, James~W Bentz, L{\'e}on Bottou, Isabelle Guyon, Yann LeCun, Cliff
  Moore, Eduard S{\"a}ckinger, and Roopak Shah. 1993.
\newblock Signature verification using a “siamese” time delay neural
  network.
\newblock \emph{International Journal of Pattern Recognition and Artificial
  Intelligence}, 7(04):669--688.

\bibitem[{Chang et~al.(2019)Chang, Felix, Chang, Yang, and
  Kumar}]{chang2019pre}
Wei-Cheng Chang, X~Yu Felix, Yin-Wen Chang, Yiming Yang, and Sanjiv Kumar.
  2019.
\newblock Pre-training tasks for embedding-based large-scale retrieval.
\newblock In \emph{International Conference on Learning Representations}.

\bibitem[{Chen et~al.(2017)Chen, Fisch, Weston, and Bordes}]{chen2017reading}
Danqi Chen, Adam Fisch, Jason Weston, and Antoine Bordes. 2017.
\newblock Reading wikipedia to answer open-domain questions.
\newblock In \emph{Proceedings of the 55th Annual Meeting of the Association
  for Computational Linguistics (Volume 1: Long Papers)}, pages 1870--1879.

\bibitem[{Devlin et~al.(2019)Devlin, Chang, Lee, and
  Toutanova}]{devlin2019bert}
Jacob Devlin, Ming-Wei Chang, Kenton Lee, and Kristina Toutanova. 2019.
\newblock Bert: Pre-training of deep bidirectional transformers for language
  understanding.
\newblock In \emph{Proceedings of the 2019 Conference of the North American
  Chapter of the Association for Computational Linguistics: Human Language
  Technologies, Volume 1 (Long and Short Papers)}, pages 4171--4186.

\bibitem[{Gururangan et~al.(2020)Gururangan, Marasovi{\'c}, Swayamdipta, Lo,
  Beltagy, Downey, and Smith}]{gururangan-etal-2020-dont}
Suchin Gururangan, Ana Marasovi{\'c}, Swabha Swayamdipta, Kyle Lo, Iz~Beltagy,
  Doug Downey, and Noah~A. Smith. 2020.
\newblock \href {https://doi.org/10.18653/v1/2020.acl-main.740} {Don{'}t stop
  pretraining: Adapt language models to domains and tasks}.
\newblock In \emph{Proceedings of the 58th Annual Meeting of the Association
  for Computational Linguistics}, pages 8342--8360, Online. Association for
  Computational Linguistics.

\bibitem[{Guu et~al.(2020)Guu, Lee, Tung, Pasupat, and Chang}]{guu2020realm}
Kelvin Guu, Kenton Lee, Zora Tung, Panupong Pasupat, and Ming-Wei Chang. 2020.
\newblock Realm: Retrieval-augmented language model pre-training.
\newblock \emph{arXiv preprint arXiv:2002.08909}.

\bibitem[{Holtzman et~al.(2020)Holtzman, Buys, Du, Forbes, and
  Choi}]{HoltzmanBDFC20}
Ari Holtzman, Jan Buys, Li~Du, Maxwell Forbes, and Yejin Choi. 2020.
\newblock \href {https://openreview.net/forum?id=rygGQyrFvH} {The curious case
  of neural text degeneration}.
\newblock In \emph{8th International Conference on Learning Representations,
  {ICLR} 2020, Addis Ababa, Ethiopia, April 26-30, 2020}. OpenReview.net.

\bibitem[{Izacard and Grave(2020)}]{izacard2020leveraging}
Gautier Izacard and Edouard Grave. 2020.
\newblock Leveraging passage retrieval with generative models for open domain
  question answering.
\newblock \emph{arXiv preprint arXiv:2007.01282}.

\bibitem[{Joshi et~al.(2017)Joshi, Choi, Weld, and
  Zettlemoyer}]{joshi2017triviaqa}
Mandar Joshi, Eunsol Choi, Daniel~S Weld, and Luke Zettlemoyer. 2017.
\newblock Triviaqa: A large scale distantly supervised challenge dataset for
  reading comprehension.
\newblock In \emph{Proceedings of the 55th Annual Meeting of the Association
  for Computational Linguistics (Volume 1: Long Papers)}, pages 1601--1611.

\bibitem[{Karpukhin et~al.(2020)Karpukhin, O{\u{g}}uz, Min, Wu, Edunov, Chen,
  and Yih}]{karpukhin2020dense}
Vladimir Karpukhin, Barlas O{\u{g}}uz, Sewon Min, Ledell Wu, Sergey Edunov,
  Danqi Chen, and Wen-tau Yih. 2020.
\newblock Dense passage retrieval for open-domain question answering.
\newblock \emph{arXiv preprint arXiv:2004.04906}.

\bibitem[{Kwiatkowski et~al.(2019)Kwiatkowski, Palomaki, Redfield, Collins,
  Parikh, Alberti, Epstein, Polosukhin, Devlin, Lee
  et~al.}]{kwiatkowski2019natural}
Tom Kwiatkowski, Jennimaria Palomaki, Olivia Redfield, Michael Collins, Ankur
  Parikh, Chris Alberti, Danielle Epstein, Illia Polosukhin, Jacob Devlin,
  Kenton Lee, et~al. 2019.
\newblock Natural questions: a benchmark for question answering research.
\newblock \emph{Transactions of the Association for Computational Linguistics},
  7:453--466.

\bibitem[{Lee et~al.(2019)Lee, Chang, and Toutanova}]{lee2019latent}
Kenton Lee, Ming-Wei Chang, and Kristina Toutanova. 2019.
\newblock Latent retrieval for weakly supervised open domain question
  answering.
\newblock In \emph{Proceedings of the 57th Annual Meeting of the Association
  for Computational Linguistics}, pages 6086--6096.

\bibitem[{Lewis et~al.(2020{\natexlab{a}})Lewis, Liu, Goyal, Ghazvininejad,
  Mohamed, Levy, Stoyanov, and Zettlemoyer}]{lewis-etal-2020-bart}
Mike Lewis, Yinhan Liu, Naman Goyal, Marjan Ghazvininejad, Abdelrahman Mohamed,
  Omer Levy, Veselin Stoyanov, and Luke Zettlemoyer. 2020{\natexlab{a}}.
\newblock {BART}: Denoising sequence-to-sequence pre-training for natural
  language generation, translation, and comprehension.
\newblock In \emph{Proceedings of the 58th Annual Meeting of the Association
  for Computational Linguistics}, pages 7871--7880, Online. Association for
  Computational Linguistics.

\bibitem[{Lewis et~al.(2020{\natexlab{b}})Lewis, Perez, Piktus, Petroni,
  Karpukhin, Goyal, K{\"u}ttler, Lewis, Yih, Rockt{\"a}schel
  et~al.}]{lewis2020retrieval}
Patrick Lewis, Ethan Perez, Aleksandara Piktus, Fabio Petroni, Vladimir
  Karpukhin, Naman Goyal, Heinrich K{\"u}ttler, Mike Lewis, Wen-tau Yih, Tim
  Rockt{\"a}schel, et~al. 2020{\natexlab{b}}.
\newblock Retrieval-augmented generation for knowledge-intensive nlp tasks.
\newblock \emph{arXiv preprint arXiv:2005.11401}.

\bibitem[{Lewis et~al.(2020{\natexlab{c}})Lewis, Stenetorp, and
  Riedel}]{lewis2020question}
Patrick Lewis, Pontus Stenetorp, and Sebastian Riedel. 2020{\natexlab{c}}.
\newblock Question and answer test-train overlap in open-domain question
  answering datasets.
\newblock \emph{arXiv preprint arXiv:2008.02637}.

\bibitem[{Miller et~al.(2016)Miller, Fisch, Dodge, Karimi, Bordes, and
  Weston}]{miller2016key}
Alexander Miller, Adam Fisch, Jesse Dodge, Amir-Hossein Karimi, Antoine Bordes,
  and Jason Weston. 2016.
\newblock Key-value memory networks for directly reading documents.
\newblock In \emph{Proceedings of the 2016 Conference on Empirical Methods in
  Natural Language Processing}, pages 1400--1409.

\bibitem[{Robertson and Zaragoza(2009)}]{robertson2009probabilistic}
Stephen Robertson and Hugo Zaragoza. 2009.
\newblock The probabilistic relevance framework: Bm25 and beyond.
\newblock \emph{Foundations and Trends in Information Retrieval},
  3(4):333--389.

\bibitem[{Salton and McGill(1986)}]{10.5555/576628}
Gerard Salton and Michael~J. McGill. 1986.
\newblock \emph{Introduction to Modern Information Retrieval}.
\newblock McGraw-Hill, Inc., USA.

\bibitem[{Shakeri et~al.(2020)Shakeri, Nogueira~dos Santos, Zhu, Ng, Nan, Wang,
  Nallapati, and Xiang}]{shakeri-etal-2020-end}
Siamak Shakeri, Cicero Nogueira~dos Santos, Henghui Zhu, Patrick Ng, Feng Nan,
  Zhiguo Wang, Ramesh Nallapati, and Bing Xiang. 2020.
\newblock \href {https://www.aclweb.org/anthology/2020.emnlp-main.439}
  {End-to-end synthetic data generation for domain adaptation of question
  answering systems}.
\newblock In \emph{Proceedings of the 2020 Conference on Empirical Methods in
  Natural Language Processing (EMNLP)}, pages 5445--5460, Online. Association
  for Computational Linguistics.

\bibitem[{Sultan et~al.(2020)Sultan, Chandel, Astudillo, and
  Castelli}]{sultan2020importance}
Md~Arafat Sultan, Shubham Chandel, Ram{\'o}n~Fernandez Astudillo, and Vittorio
  Castelli. 2020.
\newblock On the importance of diversity in question generation for qa.
\newblock In \emph{Proceedings of the 58th Annual Meeting of the Association
  for Computational Linguistics}, pages 5651--5656.

\bibitem[{Tsatsaronis et~al.(2015)Tsatsaronis, Balikas, Malakasiotis, Partalas,
  Zschunke, Alvers, Weissenborn, Krithara, Petridis, Polychronopoulos
  et~al.}]{tsatsaronis2015overview}
George Tsatsaronis, Georgios Balikas, Prodromos Malakasiotis, Ioannis Partalas,
  Matthias Zschunke, Michael~R Alvers, Dirk Weissenborn, Anastasia Krithara,
  Sergios Petridis, Dimitris Polychronopoulos, et~al. 2015.
\newblock An overview of the bioasq large-scale biomedical semantic indexing
  and question answering competition.
\newblock \emph{BMC bioinformatics}, 16(1):138.

\bibitem[{Zhang et~al.(2020)Zhang, Gangi~Reddy, Sultan, Castelli, Ferritto,
  Florian, Sarioglu~Kayi, Roukos, Sil, and Ward}]{zhang-etal-2020-multi-stage}
Rong Zhang, Revanth Gangi~Reddy, Md~Arafat Sultan, Vittorio Castelli, Anthony
  Ferritto, Radu Florian, Efsun Sarioglu~Kayi, Salim Roukos, Avi Sil, and Todd
  Ward. 2020.
\newblock \href {https://www.aclweb.org/anthology/2020.emnlp-main.440}
  {Multi-stage pre-training for low-resource domain adaptation}.
\newblock In \emph{Proceedings of the 2020 Conference on Empirical Methods in
  Natural Language Processing (EMNLP)}, pages 5461--5468, Online. Association
  for Computational Linguistics.

\end{thebibliography}
\bibliographystyle{acl_natbib}

\clearpage

\section*{Appendix}

\subsection*{Hyperparameters}

In this section, we share the hyperparameters details for our experiments. Table \ref{tab:hyp_gen} gives the hyperparameters for training the generator and Table \ref{tab:hyp_ir} lists the hyperparameters for pre-training and finetuning the neural IR model. 

Our BM25 baseline is based on Lucene\footnote{\href{https://lucene.apache.org}{https://lucene.apache.org}}  Implementation. BM25 parameters $b=0.75$ (document length normalization) and $k_1=1.2$ (term frequency scaling) worked best.

\begin{table}[ht]
\centering
\begin{tabular}{c|c}
Hyperparameter & Value \\
\hline
Learning rate  & 3e-5\\
Epochs  & 3\\
Batch size  & 24 \\
Max Sequence length & 1024\\
\end{tabular}
\caption{Hyperparameter settings during training the synthetic example generator (BART) using data from NQ.}
\label{tab:hyp_gen}
\end{table}

\begin{table}[ht]
\centering
\small
\begin{tabular}{c|c|c}
Hyperparameter & Pre-training & Finetuning \\
\hline
Learning rate  & 1e-5 & 1e-5\\
Epochs  & 6 & 20\\
Batch size  & 1024 & 128 \\
Gradient accumulation steps & 8 & 1 \\
Max Sequence length & 256 & 256\\
\end{tabular}
\caption{Hyperparameter settings for the neural IR model during pre-training on synthetic data and finetuning on NQ.}
\label{tab:hyp_ir}
\end{table}

\subsection*{How Many Synthetic Examples do We Need?}

To analyze how much synthetic data is required, we experiment with pre-training using 1M and 4M synthetic examples while keeping the number of training updates fixed. 
As Table \ref{tab:syn_exm_analysis_app} shows, we don't see improvements from using more examples beyond 2M.

\begin{table}[h]
\centering
\begin{tabular}{l|c|c|c}
Model & Top-10 & Top-20 & Top-100 \\
\hline
DPR & 73.6 & 78.1 & 85.0 \\
AugDPR-1M & 74.4 & 79.2 & 85.5 \\
AugDPR-2M & 74.8 & 79.7 & 85.9\\
AugDPR-4M & 74.6  & 79.1 & 85.9\\
\end{tabular}
\caption{Retrieval accuracy on the Natural Questions dev set with varying number of synthetic examples (1M vs 2M vs 4M) during pre-training.}
\label{tab:syn_exm_analysis_app}
\end{table}

\end{document}